# Augmenting High-dimensional Nonlinear Optimization with Conditional GANs

Pouya Rezazadeh Kalehbasti
Stanford University
Stanford, CA 94305
USA
pouyar@stanford.edu

Michael D. Lepech
Stanford University
Stanford, CA 94305
USA
mlepech@stanford.edu

Samarpreet Singh Pandher
Stanford University
Stanford, CA 94305
USA
samar89@stanford.edu

## ABSTRACT

Many mathematical optimization algorithms fail to sufficiently explore the solution space of high-dimensional nonlinear optimization problems due to the curse of dimensionality. This paper proposes generative models as a complement to optimization algorithms to improve performance in problems with high dimensionality. To demonstrate this method, a conditional generative adversarial network (C-GAN) is used to augment the solutions produced by a genetic algorithm (GA) for a 311-dimensional nonconvex multi-objective mixed-integer nonlinear optimization. The C-GAN, composed of two networks with three fully connected hidden layers, is trained on solutions generated by GA, and then given sets of desired labels (i.e., objective function values), generates complementary solutions corresponding to those labels. Six experiments are conducted to evaluate the capabilities of the proposed method. The generated complementary solutions are compared to the original solutions in terms of optimality and diversity. The generative model generates solutions with objective functions up to 100% better, and with hypervolumes up to 100% higher, than the original solutions. These findings show that a C-GAN with even a simple training approach and architecture can, with a much shorter runtime, highly improve the diversity and optimality of solutions found by an optimization algorithm for a high-dimensional nonlinear optimization problem. [Link to GitHub repository: https://github.com/PouyaREZ/GAN_GA]

## CCS CONCEPTS

• **Computing methodologies** → **Neural networks** • **Theory of computation** → **Evolutionary algorithms** • Operations Research → Multi-criterion optimization and decision-making

## KEYWORDS

Conditional Generative Adversarial Nets, Genetic Algorithm, Multi Objective Optimization, Sustainable Energy Systems

## 1 INTRODUCTION

Mathematical optimization faces challenges when solving highly non-linear and high-dimensional optimization problems [13,23]. This is because the solution space for these problems is so vast that the optimization algorithm fails to properly explore the optimal solutions of the entire space in a reasonable amount of time [13,23]. To solve this issue, this paper proposes using conditional generative adversarial networks (C-GANs) to learn the underlying distribution of the solutions generated by the optimization algorithm, and then generating unseen, more optimized solutions to the original optimization problem using the generative model. C-GAN consists of two adversarial models, a generator and a discriminator [30]. The generator learns the data distribution of the input solutions to generate new solutions, and the discriminator learns to detect if a solution belongs to the input data distribution or not. The adversarial training as well as conditioning on the data labels enables C-GAN to generate unseen solutions for given desired labels [30].

The proposed method is tested on a nonconvex multi-objective mixed-integer nonlinear program (MINLP), which is solved using a genetic algorithm (GA). This optimization problem concerns the sustainable design of the buildings, energy plant, and energy distribution network in an urban district by minimizing the life-cycle cost (LCC) and greenhouse gas emissions (GHG) and maximizing the walkability (WalkScore) [39] of the district. A C-GAN is used to generate complementary solutions for the optimization problem based on the solutions that are output by the optimization algorithm. The problem of training the C-GAN on the solutions of the optimization problem is handled as one of multi-variate multiple regression [22], where the features (independent variables) of the training set are the integer inputs of the optimization problem, and the labels (dependent variables) are the real-valued objective functions.

### 1.1 Background

#### 1.1.1 Nonconvex Multi-objective MINLP

The optimization problem this study tackles is one of nonconvex MINLP, an NP-complete problem where the decision variables are a mix of integers and continuous variables and nonlinear objective functions and/or constraints [10]. This type of optimization problem occurs, for example, in optimizing portfolios, chemical batch processing, and designing energy, water, or gas networks [9,10,46]. Several exact solutions (e.g. branch-and-bound with convexification [38], branch-and-reduce [36], and $\alpha$-branch-and-bound [2]), as well as heuristic methods [9] (e.g. Tabu search [16], particle-swarm optimization [41], and genetic algorithm [42]) have been devised for solving nonconvex MINLPs.

The MINLP problem studied in this paper has multiple objectives. This adds another layer of complexity to finding optimal solutions to the problem. In such problems, the optimization algorithm tries to optimize multiple, potentially conflicting, objective functions at the same time [29]. The general form of multi-objective optimization is as follows:

$$Minimize\ F(x) = \{f_1(x), \ldots, f_O(x)\};\ subject\ to\ x \in X \quad (1)$$

where $x$ is the vector of decision variables, $f_i(x)$ is one of the $O$ objective functions, and $X$ is the set of all admissible solutions. In practice, evolutionary algorithms (notably, NSGA-II [14], SPEA2 [47], and SMS-EMOA [8]), particle-swarm optimization [34], and ant colony optimization [5] are some of the most popular methods for solving nonlinear, multi-objective optimization problems [15].

Several factors make the solution space of the optimization problem studied in this paper very hard to explore with routine optimization algorithms. The evaluation (aka fitness, objective) function, which maps the permissible solutions to objective functions, is highly nonlinear. The overall solution space has 311 variables, i.e. it is 311-dimensional, which are a mix of integer and continuous values. Finally, the optimization has a two-level scheme. Thus, generative models can be used in this case to further explore the solution space and to find more optimized solutions to the problem [20,23]. Generative models can learn the underlying distribution of the solutions generated by the optimization algorithms, and then generate, potentially, better solutions to the optimization problem.

*1.1.2    Generative Adversarial Networks*

Among different generative models, GANs [19] have recently garnered attention among researchers as robust models for learning highly complex underlying distributions of data samples. These models have shown promising results when applied to problems from generic and medical image generation [4,21] to time-series analysis [43], to regression [1,12,27], and image manipulations [24,26,44].

A GAN encompasses two adversarial models which are trained at the same time. A generator, $G$, attempts to learn the data distribution and generate samples belonging to that distribution. A discriminator, $D$, attempts to distinguish real samples from those fabricated by $G$ [30]. Conditional GANs are a special type of GANs where, (i) $G$ builds a mapping function from a prior noise distribution ($p_z$) to the data distribution ($p_x$) as $G(z|y;\theta_g)$ conditioned on the label ($y$), and (ii) the discriminator builds a function $D(x|y;\theta_d)$ conditioned on the label ($y$) which outputs the probability of $x$ belonging to the same distribution as the training data [30]. $\theta_g$ and $\theta_d$ are the set of parameters for $G$ and $D$, respectively. These two models engage in a two-player minimax game with an objective function described by Equation (2) [30]:

$$\min_G \max_D V(D,G) = E_{x \sim p_x(x)}[\log D(x|y)] + E_{z \sim p_z(z)}[\log(1 - D(G(z|y)))] \quad (2)$$

where $x$ is the feature vector (i.e. actual data point) and $z$ is the noise vector.

*1.1.3    GANs and Optimization*

Given their potential for learning complex data distributions, some recent studies have applied GANs to some optimization problems. Guo et al. [20] and He et al. [23] used GANs to improve the offspring generation of a genetic algorithm (GA) by training GANs on the offspring produced by a GA at each generation, then randomly generating more offspring, and finally selecting the desired offspring from these two sets. Guo et al. and He et al. both observed improvements in offspring generation by applying GANs. Zhao and You [45] used GANs to generate uncertain parameters in chance constrained programming, where the parameters generated by the GAN were fed to an optimization algorithm. Zhao and You also reported improved results from the uncertain parameters generated by a GAN. Rawat and Shen [33] attempted image generation using Wasserstein GANs (WGAN) for structural topology optimization. They trained a WGAN on images of the optimal solutions generated by an optimization algorithm and tried to generate images of more diverse optimal solutions using the WGAN. The results reported by Rawat and Shen were not promising compared to the results of Guo et al., He et al., and Zhao and You.

*1.1.4    Contributions*

Research to date on applying GANs to optimization has focused on random generation rather than targeted generation of data points with desired labels. This paper treats the task of generating solutions to an optimization problem as a regression, and uses C-GANs to generate solutions with desired labels (objective function values) rather than relying only on random generation of solutions. Related works have seen promising results from applying C-GANs to regression [1,12,27,40].

Other studies in the literature that apply GANs to GAs have trained GANs only on the limited data generated during off-spring production of GAs. However, this paper trains a GAN on the final results of the GA, thereby improving the GAN with a larger training set and saving significant runtime by training the GAN model only once. The contributions of this work include, (i) a new method for augmenting traditional optimization for highly complex optimization problems, (ii) the first application in the literature of C-GANs to multivariate multiple regression, and (iii) the first direct application in the literature of C-GANs to mathematical optimization.

## 2    Methods

This paper uses a C-GAN to complement the performance of a genetic algorithm on a high-dimensional nonlinear optimization. The C-GAN trains on the results of the GA, and then generates more diverse and more optimized solutions to the optimization problem than the GA has identified. The generated solutions that satisfy the constraints of the original optimization problem (i.e. admissible solutions) are kept as the complementary solutions. Figure 1 shows the proposed algorithm.

Augmenting High-dimensional Nonlinear Optimization with Conditional GANs

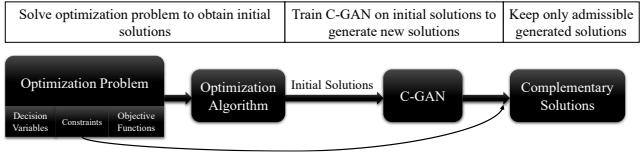

**Figure 1: Overview of the proposed algorithm for using a C-GAN to complement an optimization algorithm on a high-dimensional nonlinear optimization**

## 2.1 Optimization Problem

The non-convex multi-objective MINLP studied in this paper addresses the sustainable design of the buildings, energy plant, and energy distribution network in an urban neighborhood (aka district) by minimizing the life-cycle cost (LCC) and greenhouse gas emissions (GHG) and maximizing the walkability (WalkScore) [39] of the neighborhood. The neighborhood considered is a four-node grid where buildings 'can' occupy three nodes out of the four and a central energy plant occupies one of the remaining nodes. Figure 2 shows the sketch of a sample grid — the number of buildings in the grid can vary from one to three, and the buildings can be located at any of the four nodes. The four nodes are located at coordinates (0,0), (0,100), (100,0), and (200,100), in meters.

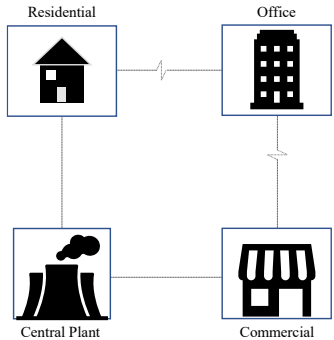

**Figure 2: Sample configuration of the neighborhood grid designed by the optimization. The number of buildings, their locations, and the location of the central plant are arbitrary.**

The optimization is done in two layers. In the primary optimization, the algorithm chooses the location and the type of the buildings from four types. Building types can be residential, commercial, office, or hospitality. The algorithm also chooses the type of the combined heating and power (CHP) engine from six predefined models to provide heating and electricity, and the type of the chiller from three predefined models to provide cooling to the buildings. The CHP engine and the chiller are collocated at the central plant which occupies a single node in the grid. The following describes the optimization problem.

*Objective Functions*
 Minimize life-cycle cost
 Minimize greenhouse gas emissions
 Maximize walkability

*Decision Variables*
 Types of buildings at nodes (four integers $\in [1,4]$)
 Type of CHP engine (integer $\in [1,6]$)
 Type of chiller (integer $\in [1,3]$)
 Temperature of hot water (integer $\in [50,95]$)
 Reset temperature of hot water in summer (integer $\in [0,10]$)
 Temperature of cold water (integer $\in [1,8]$)
 Reset temperature of cold water in winter (integer $\in [0,3]$)

*Constraints*
 At least one node must be occupied by a building
 At most three nodes must be occupied by buildings
 Decision variables must be within the defined ranges

This over-arching optimization problem is a multi-objective integer non-linear programming (MO-INLP) based on the work by Best et al. [7]. This optimization is solved with the genetic algorithm NSGA-II [14] using Deap 1.3 library [18] in Python 3.8 [35]. The GA is run over 512 generations of 128 individuals each, generating 65,610 solutions. The hyperparameters of the GA are set following Best et al. [7], with mutation probability, cross probability, and $\eta$ set to 0.05, 0.75 and 2.5, respectively.

Given the types and locations of the buildings and engines, a secondary optimization algorithm selects the types and locations of the pipes (out of five options) that transfer the hot and cold water from the plant to the buildings such that the life-cycle cost of the heating and cooling network (including construction and operation of the network) is minimized. The optimization algorithm for this sub-routine is based on the method introduced by Best et al. [6]. This optimization problem is a mixed-integer linear programming (MILP) and is solved with Gurobi Optimizer 9.1.0 [32] using the PuLP 2.3 [31] library in Python.

## 2.2 Conditional Generative Adversarial Network

C-GANs can augment the optimization problem defined here, as they can generate features (i.e., neighborhood characteristics) given the labels (i.e., the objective function values). Hence, by feeding the desired labels to the C-GAN, the model can generate the input features corresponding to those labels. The C-GAN used in this work comprises a fully-connected generator and a fully-connected discriminator. The base architecture of the network was taken from the "CGAN" model in the GitHub repository "Keras-GAN" [28] which itself is a simplified version of the model proposed by Mirza and Osindero [30].

Following manual tuning of hyperparameters [detailed in Appendix A], the architectures of the generator and discriminator were set as shown in Figure 3. A more complicated architecture is devised for the generator as it has a harder task to accomplish than the discriminator. The Leaky ReLU layers have an alpha of 0.2. The dropout layers have a probability of 0.4. The Batch Normalization has a momentum of 0.8 and $\epsilon$ of 0.001. The 3-dimensional noise, denoted by $\boldsymbol{Z}$ in Figure 3(a), is generated from the gaussian distribution $N(0,1)$. Adam Optimizer [25] with a learning rate of 0.0002 and $(\beta_1, \beta_2)$ of (0.5, 0.999) is used to minimize the loss function "binary cross-entropy" for both the discriminator and the generator.

The noise and the label, as well as the input features and the label, are concatenated together and fed into, respectively, the generator and the discriminator (Figure 3).

### 2.3 Experiments

Different subsets of the 65,610 solutions to the optimization problem, described in section 2.1, are used to perform six experiments to inspect the capabilities of the proposed method. In these experiments, the C-GAN is trained on different subsets of the original solutions which simulate situations where the optimization algorithm has produced solutions with desirable or undesirable objective function values. The solutions that C-GAN generates are then compared, in terms of different objective function values, with their respective training sets.

Table 1 describes the six experiments conducted in this paper, their training sets, and their goals. In the first three experiments, the C-GAN is trained on the worst half of the initial solutions in terms of *each objective function*. These three experiments measure how well C-GAN can generate solutions with, respectively, improved GHGs, LCCs, and WalkScores compared to a training set composed of adversely selected solutions in terms of each objective function. In *WorstHalfAll*, the same procedure is followed for a training set composed of solutions with *all three objective functions* in the worst half of the objective function values in the initial solutions. The generated solutions are then compared in terms of all objective functions with the training set.

In *BestHalfAll* (Table 1), the C-GAN is trained on solutions with *all three objective functions* in the best half of the objective function values in the initial solutions. This experiment measures if the C-GAN can generate solutions with better objective function values than a training set of elite solutions found by the optimization algorithm. In *FullData*, the C-GAN is trained on the entire 65,610 initial solutions. This final experiment measures the performance of C-GAN in generating solutions with better objective function values compared to the initial solutions. Table 2 lists the size of the training set for each experiment under the second column, denoted '**#Train**.'

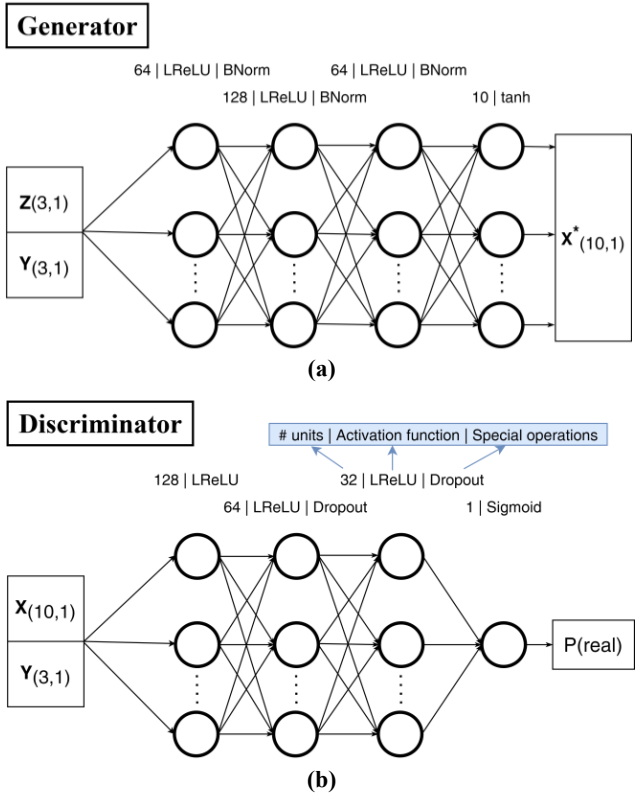

Figure 3: The architectures of (a) the generator and (b) the discriminator networks. [BatchNorm: Batch Normalization, LReLU: Leaky ReLU]

The outputs of the overall optimization algorithm (described in the previous subsection) are the inputs to the C-GAN. Hence, the C-GAN network sees the following ten integers as the input features of the training dataset.
- Four integers indicate building type selections for the four grid nodes [Ranging from 0 to 5, with 0 indicating no building built at a node, 1 to 4 indicating the building type, and 5 indicating the central plant built at a node].
- An integer indicates CHP engine type selection for the central plant [ranging from 1 to 6]
- An integer indicates chiller type selection for the central plant [ranging from 1 to 3]
- Four integers indicate the temperature of the heating water supply and its reset value for the summer, and the temperature of the cooling water supply and its reset value for the winter.

As labels, C-GAN sees the following three continuous variables.
- Life-cycle cost of the neighborhood (within $(-\infty, +\infty)$, in $ per m$^2$ area of the buildings)
- Life-cycle emissions of the neighborhood (within $[0, +\infty)$, in ton of $CO_{2eq}$ per m$^2$ area of the buildings)
- WalkScore of the neighborhood (within $[0, 15]$).

Table 1: Training sets and objectives of experiments

| Experiment Name | Training Set: Solutions satisfying… | Goal: Generate diverse solutions with improved… |
| --- | --- | --- |
| *WorstHalfGHG* | $GHG \geq GHG_{median}^*$ (I) | GHGs |
| *WorstHalfLCC* | $LCC \geq LCC_{median}$ (II) | LCCs |
| *WorstHalfWalkScore* | $WalkScore \leq WalkScore_{median}$ (III) | WalkScores |
| *WorstHalfAll* | (I) & (II) & (III) | GHG, LCC, and WalkScore |
| *BestHalfAll* | ~(I) & ~(II) & ~(III) | GHG, LCC, and WalkScore |
| *FullData* | entire dataset | GHG, LCC, and WalkScore |

* *median* subscript indicates the median value of the entire 65,610 solutions

In all the experiments, the C-GAN is expected to ideally generate more diverse and more optimized solutions than those on which it is trained. In this way, the trained C-GAN complements the performance of the optimization algorithm. To assess the improvement in diversity (a.k.a. distribution [3]) and optimality of the solutions, two metrics are used. These metrics are, (i) the optimal values of the objective functions among the generated solutions, and (ii) the hypervolume (HV) [17] of the generated Pareto front. HV can also measure the convergence of the Pareto front [3]. These metrics are calculated and compared with those of the original dataset.

At set intervals (e.g., every 100 iterations) during the training procedure, custom-built labels are fed into the generator to produce a set of generated inputs given those labels (Appendix B details these custom labels for each experiment). A subset of these generated inputs is selected at the end of the training as the candidate complementary solutions. These inputs are selected from stages at which either the discriminator achieves high accuracy identifying actual solutions or has a low loss [indicating a strong discriminator challenging the generator to try its best], or when the discriminator has a low accuracy identifying the generated solutions or the generator has a low loss [indicating a strong generator]. The quality of the generated solutions is expected to be higher under these circumstances. These candidate solutions are ultimately aggregated for each experiment and vetted using the original evaluation function to check their admissibility (i.e., complying with the constraints of the original optimization problem and not being duplicates of the original solutions) and to calculate the exact labels for the admissible solutions.

For every experiment, the results of two selected runs are combined; the two runs include one long run over roughly 155 epochs of training and a short run over a small number of epochs. The long run is conducted to achieve a fully trained and stable generative model, while the short run is conducted to use early-stopping to avoid potential overfitting of the model [11]. The number of iterations for the short runs is customized for each experiment, such that the training curves (both loss and accuracy) relatively stabilize when the model is trained up to that iteration. Combining the results of these two training regimes (i.e., the short run and the long run) leads to higher values for both the hypervolume and the objective functions compared to the values of these metrics calculated for the results processed separately.

The supplementary materials include a Jupyter Notebook (along with its dependencies) that can replicate all the experiments.

## 3 RESULTS AND DISCUSSION

Figure 4 shows the plots of accuracy and loss as a function of iteration number for both the generator and the discriminator from the short run in the BestHalfAll experiment. The figure shows that from iteration 500 onwards, the discriminator and generator come to a state of equilibrium, which is a desirable outcome when training a GAN [37]. Tuning the hyperparameters (including the batch-size and the number of iterations) led to achieving the same equilibrium for most of the other experiments done in this study.

On a single core of a 2.2 GHz Intel Xeon, the optimization algorithm takes 2096 minutes to obtain the 65,610 solutions, while training the C-GAN and generating the complementary solutions (combining the short and long run) takes from 8 to 59 minutes proportional to the number of training iterations. This is less than 3% of the runtime of the optimization algorithm.

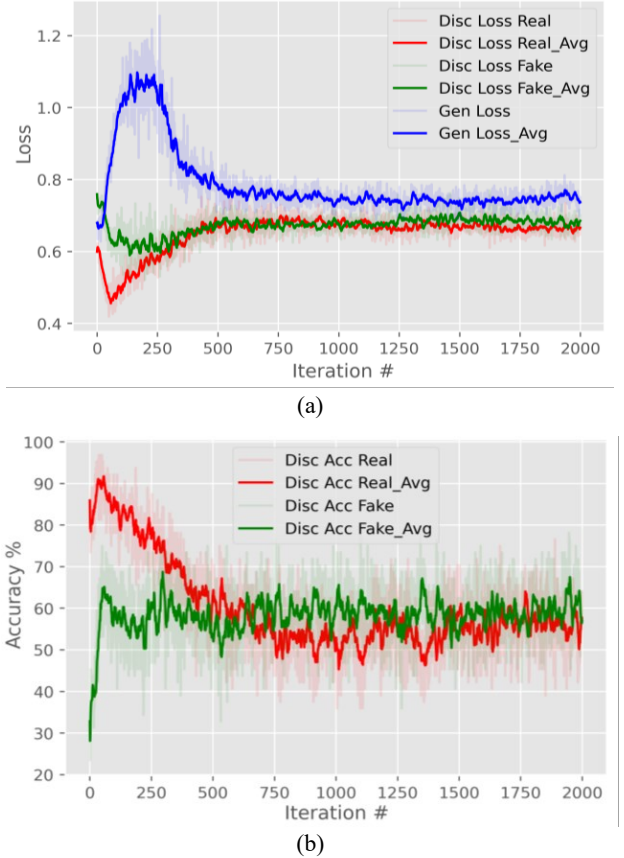

Figure 4: Plots of training loss (a) and accuracy (b) from the short run of experiment BestHalfAll. The sharper curves are the running average of the transparent curves, with a window size of 10. [Acc: accuracy, Avg: running average, Disc: discriminator, Gen: generator]

Table 2 contains (from left to right) the number of training and generated datasets (**#Train** and **#Gen**), the admissibility ratio of the generated solutions (**Admissible**), the minimum value of GHG (**Min GHG (T-CO$_2$/m$^2$)**), the minimum value of LCC (**Min LCC ($/m$^2$)**), and the maximum value of WalkScore (**Max WalkScore**) observed among the training data (**Train**) and the generated data (**Gen**) as well as the percent improvement in these values when comparing the generated data with the training data (**Improved**). The last column-group from the left in Table 2 (**Hypervolume**) shows the values of HV for the training set and the generated set, and the percentage of improvement in this value when comparing the generated with the original data. To calculate the HV, the three objective functions of both the training and generated solutions are scaled to a range from 0 to 1 using min-

max scaling. 0 and 1 correspond to, respectively, the most and the least desirable values of each objective function in the training set. Point (1,1,1) is taken as the reference point [17] for the HV.

The admissibility ratio (**Admissible**) in Table 2 indicates the level of effort required by the generator to learn the underlying distribution of the solutions and to generate solutions with a higher quality than the training data. The generator faces the greatest challenges when generating new solutions based on the WorstHalfWalkScore dataset, as the generated solutions in this experiment have the lowest admissibility ratio (~19%) among all experiments. As mentioned in Table 1, the objective of this experiment is to generate solutions with improved WalkScores. However, as the **Train** column under **Max WalkScore** group of columns shows, all datapoints in the training set of this experiment have a WalkScore of 0, which has made it difficult for the generator to create solutions with non-zero WalkScores. Despite this challenge, the generator generates solutions with the maximum possible WalkScore values, while also greatly improving the hypervolume of the Pareto front compared to the input dataset (look under **Hypervolume** column-group).

Table 2 also shows that the generator generates solutions with a minimum GHG of 0.76 T-$CO_2$/$m^2$ for the first and the last three experiments, improving the minimum GHG observed in the input datasets by up to 21%. This minimum GHG value is equal to the lowest GHG value seen among the entire training dataset.

Similarly, the generator produces solutions with WalkScores equal to the maximum WalkScore observed in the original solution set (i.e., 15.0) in the last four experiments. The C-GAN improves the maximum value of WalkScore by up to 100%, from a score of 0.0 to a score of 15.0.

The C-GAN also generates solutions with minimum LCCs up to 79% lower than those of the input datasets. Similar to the case of minimum GHG, these solutions have a minimum LCC close to the minimum value observed across the entire original solution set (i.e., -7680 \$/$m^2$). However, the generative model cannot improve the minimum value of LCC in the FullData experiment, which indicates that the original optimization algorithm has discovered a solution with a value close to the global optimum LCC.

**Table 2: Size, optimal value of objective functions, and hypervolume of the training and generated solutions in the six experiments [Train: training data, Gen: generated data]**

| Experiment | Metadata | | | Min GHG (T-$CO_2$/$m^2$) | | | Min LCC (\$/$m^2$) | | | Max WalkScore | | | Hypervolume | | |
|---|---|---|---|---|---|---|---|---|---|---|---|---|---|---|---|
| | #Train | #Gen | Admissible | Train | Gen | Improved | Train | Gen | Improved | Train | Gen | Improved[*] | Train | Gen | Improved[*] |
| WorstHalfGHG | 33438 | 875 | 79.1% | 0.83 | 0.76 | 9.0% | | | | | | | 0.997 | 0.997 | 0.07% |
| WorstHalfLCC | 32796 | 875 | 77.9% | | | | -4281 | -7648 | 78.7% | | | | 0.990 | 0.997 | 0.71% |
| WorstHalf–WalkScore | 40073 | 875 | 18.5% | | | | | | | 0.0 | 15.0 | 100.0% | 0.000 | 0.995 | 100.00% |
| WorstHalfAll | 5145 | 2750 | 41.1% | 0.96 | 0.76 | 21.0% | -4281 | -7464 | 74.4% | 0.0 | 15.0 | 100.0% | 0.000 | 0.996 | 100.00% |
| BestHalfAll | 7989 | 5000 | 81.2% | 0.83 | 0.76 | 8.4% | -4648 | -7478 | 60.9% | 9.5 | 15.0 | 57.9% | 0.631 | 0.997 | 57.87% |
| FullData | 65610 | 4000 | 65.4% | 0.76 | 0.76 | 0.0% | -7680 | -7622 | -0.8% | 15.0 | 15.0 | 0.0% | 0.997 | 0.997 | 0.05% |

[*] **Improved** is calculated based on $\frac{Gen - Train}{Gen} \times 100$ when **Train** is zero, and $\frac{Gen - Train}{Train} \times 100$ otherwise.

Table 2 shows that the generative model improves the hypervolumes of all the experiments by ~0.1% to 100.0%, which shows that compared to the input dataset, the solutions generated by the C-GAN have higher spread and convergence to the optimal Pareto front. Figure 5 shows the scatter plots of GHG vs LCC and WalkScore vs LCC for the BestHalfAll experiment. Figure 5(b) and Figure 5(d) clearly show the power of the generative model in exploring the unseen solutions to the problem, and in expanding the Pareto front.

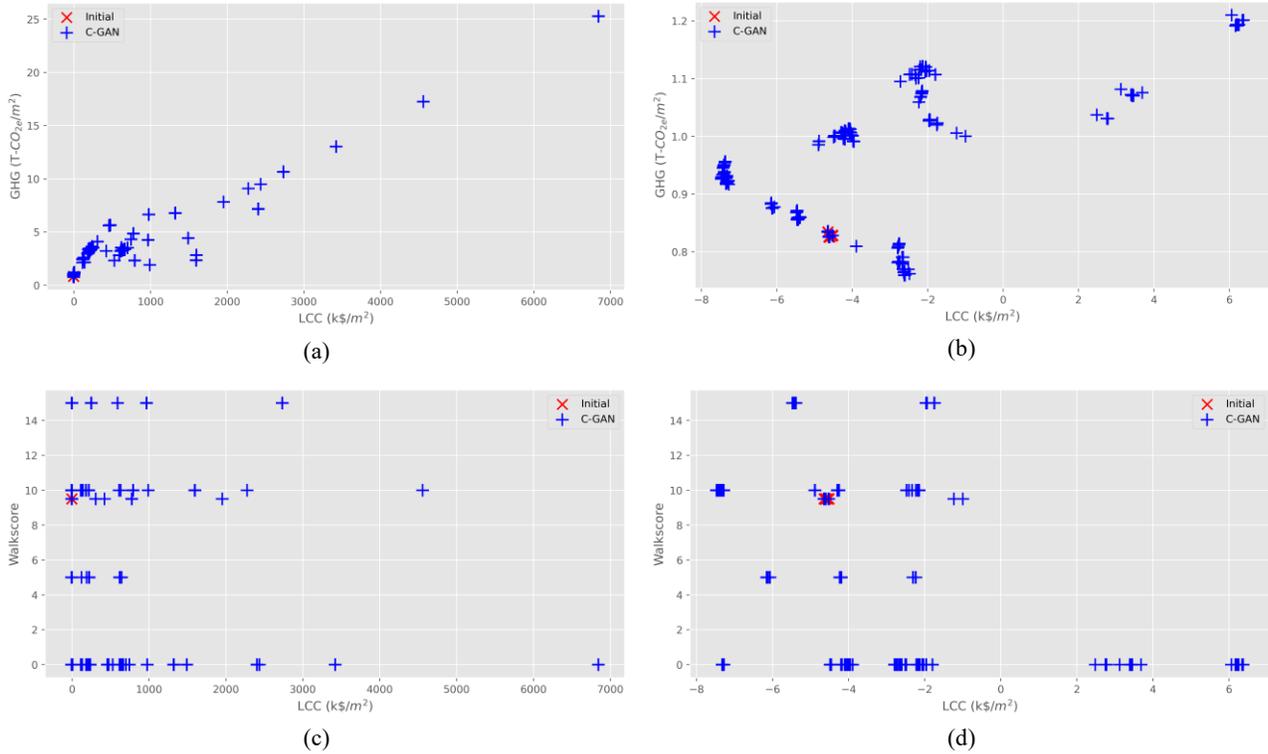

Figure 5: (a, b) GHG vs LCC and (c, d) WalkScore vs LCC plots for BestHalfAll experiment. Red '×' marks indicate the 7,989 solutions in the training set, and blue '+' marks show the 5,000 solutions in the generated set. Figures (b) and (d) show only those solutions with LCC ≤ 10 k\$/m$^2$.

## 4 CONCLUSIONS AND FUTURE WORK

This work proposes to use generative models to augment the results of optimization algorithms for high-dimensional nonlinear optimization problems. To showcase the proposed method, the solutions output by a genetic algorithm (GA) for a multi-objective mixed-integer nonlinear programming are used as the training set of a conditional generative adversarial network (C-GAN). The generator and discriminator of the C-GAN have three fully-connected hidden layers. Six different experiments are designed to test the abilities of the C-GAN to generate more optimized and more diverse solutions. Two metrics are used to evaluate the performance of the generative model in the experiments. These metrics are, (i) the best values of the objective functions among the generated solutions, and (ii) the hypervolume of the generated solutions. In both the experiments where a single objective is the target to be improved, and where all three objectives are the targets, the model successfully improves the best values of the objective functions among the generated solutions by up to 100% compared to the original solutions. The model also generates solutions with hypervolumes up to 100% higher than those of the original solutions. The C-GAN achieves these in less than 3% of the time needed to run the original optimization method. These results speak to the promise of using generative models, specifically C-GANs, for improving the performance of optimization algorithms, like genetic algorithms, for high-dimensional optimization. This paper also demonstrates that C-GANs, even with simple architectures and small training iterations on low-quality solutions, can significantly improve the results of complex optimization problems.

This paper uses the final solutions produced by a GA for an optimization problem to train the C-GAN. Future work can run the generative model during the time the GA is running. For instance, this process can be repeated: The C-GAN can be trained on the results of $x$ generations of the GA, and then let the GA process over another m generations the best solutions among those generated by the C-GAN and the original solutions. Such an approach might yield higher-quality solutions at the cost of increased runtime and increased complexity of the overall procedure for generating complementary solutions.

In this work, the evaluation function of the genetic algorithm can tractably process hundreds of individuals in a reasonable amount of time. For optimization problems where the evaluation function is too costly to run, a simpler model (i.e., a surrogate) would be necessary to learn the evaluation function and to replace it. This can be an interesting extension to the work done in this paper.

Here, the inputs to the discriminator and the generator of the C-GAN are simple concatenations of the label and the features or the noise. Future research can investigate higher order interactions of these inputs to improve the performance of the model.

For simplicity in this paper, a process similar to a grid search is used to create the custom labels for generating the complementary solutions. Other approaches, like random search or Latin Hypercube Sampling, may help improve the quality of the generated solutions. Such an approach would be a worthwhile extension to the proposed method. Similarly, this paper uses pseudo-grid-search manual hyperparameter tuning. Future work can try different methods for tuning the hyperparameters of the C-GAN.

Follow-up work should also apply the proposed method to different types of complex optimization problems, including combinatorial optimization, other mixed-integer nonlinear problems, etc. These efforts will help identify, (i) the most suitable optimization problems to apply this approach, and (ii) other generative models (e.g., GANs, DGANs) that are suitable to use with this proposed method.

## A  Hyperparameters Tested for C-GAN

Some of the values tested for hyperparameters are as follows (the finally selected values are emboldened):

- Architectures tried for the discriminator:
  64-32-16; **128-64-32;** 256-256-128; 512-256-128; 512-512-512
- Architectures tried for the generator:
  32-64-32; **64-128-64;** 128-256-128; 256-512-256
- Learning rates tried with Adam: 0.01, 0.001, 0.0001, **0.0002**
- $\beta_1, \beta_2$ tried with Adam:
  (0.9, 0.999), (0.9, 0.9), (0.5, 0.9), **(0.5, 0.999)**
- Batch sizes: 32, **64**, 128
- Latent dimension of the input noise: 1, 2, **3**, 11, 22
- Number of iterations (1 epoch ≡ training over the entire training set once):
  Short runs: 300, 500, **800**, 1000, **2000**
  Long runs: 10000, 50 epochs, 100 epochs, **155 epochs**

## B  Range of Custom Labels Fed to Generator

Table 3 summarizes the ranges of custom labels, based on [-1,1] normalized labels, fed to the generator to synthesize the complementary solutions. Depending on the experiment, the custom labels urge the generator to target desirable values of one or all three objective functions. For example in WorstHalfGHG, the second component of the label, which corresponds to the GHG of the solution, is set between -1.0 and -1.2 so that the generator aims for generating solutions with GHGs less than the lowest GHG seen in the training set. A triple-nested for-loop allowed each value in the range defined for each objective function to be combined with each value in the ranges of the other two objective functions. The total number of unique labels created this way was equal to 125 for the first three experiments, and equal to 500 for the last three.

Table 3: Ranges of labels used with the generator to create the complementary solutions

| Experiment | Range of Label (based on [-1,1] normalized labels) | | |
|---|---|---|---|
| | LCC | GHG | WalkScore |
| WorstHalfGHG | [-1.0, 1.0, 0.5]* | [-1.0, -1.2, 0.05] | [-1.0, 1.0, 0.5] |
| WorstHalfLCC | [-1.0, -1.2, 0.1] | [-1.0, 1.0, 0.5] | [-1.0, 1.0, 0.5] |
| WorstHalf—WalkScore | [-1.0, 1.0, 0.5] | [-1.0, 1.0, 0.5] | [1.0, 1.2, 0.1] |
| WorstHalfAll[+] | [-1.0, -1.2, 0.1] | [-1.0, -1.2, 0.05] | [1.0, 1.2, 0.1] |
| BestHalfAll[+] | [-1.0, -1.2, 0.1] | [-1.0, -1.2, 0.05] | [1.0, 1.2, 0.1] |
| FullData[+] | [-1.0, -1.2, 0.1] | [-1.0, -1.2, 0.05] | [1.0, 1.2, 0.1] |

* Format of the ranges: [start (inclusive), end (inclusive), size of step]

[+] For the last three experiments, three sets of labels identical to those of the first three experiments (WorstHalfGHG, WorstHalfLCC, and WorstHalfWalkscore) were also fed to the generator.

## ACKNOWLEDGMENTS

This material is based on work supported by the Leavell Fellowship on Sustainable Built Environment.